\documentclass[10pt,twocolumn,letterpaper]{article}

\usepackage{cvpr}
\usepackage{times}
\usepackage{epsfig}
\usepackage{graphicx}
\usepackage{amsmath}
\usepackage{amssymb}
\usepackage[normalem]{ulem}
\usepackage{afterpage}
\usepackage{booktabs}
\usepackage{color}
\usepackage{verbatim}
\usepackage{soul}
\usepackage{xspace}
\usepackage{comment}
\usepackage{array}
\usepackage{multirow}
\usepackage{url}
\usepackage{epigraph}

\cvprfinalcopy

\makeatletter
\DeclareRobustCommand\onedot{\futurelet\@let@token\@onedot}
\def\@onedot{\ifx\@let@token.\else.\null\fi\xspace}

\def\eg{\emph{e.g}\onedot}

\def\iid{\emph{i.i.d}\onedot}
\makeatother
\newcommand{\thickhline}{%
    \noalign {\ifnum 0=`}\fi \hrule height 1pt
    \futurelet \reserved@a \@xhline
}

\newcolumntype{"}{@{\hskip\tabcolsep\vrule width 1.5pt\hskip\tabcolsep}}

\ifcvprfinal\fi

\begin{document}
\let\svthefootnote\thefootnote

\title{PixelNet: Towards a General Pixel-Level Architecture} 
\author{Aayush Bansal$^{1}$*  \quad     \quad Xinlei Chen$^{1}$* \quad \quad   Bryan Russell$^{2}$  \quad  \quad  Abhinav Gupta$^{1}$  \quad  \quad  Deva Ramanan$^{1}$ \\
$^1$Carnegie Mellon University \quad \quad $^2$Adobe Research \\
{\tt\small{\url{http://www.cs.cmu.edu/~aayushb/pixelNet/}}}
}

\maketitle

\begin{abstract}{\let\thefootnote\relax\footnote{* indicates equal contribution; first two authors listed in alphabetical order.}\addtocounter{footnote}{-1}\let\thefootnote\svthefootnote}We explore architectures for general pixel-level prediction problems, from low-level edge detection to mid-level surface normal estimation~\cite{Bansal16} to high-level semantic segmentation. Convolutional predictors, such as the fully-convolutional network (FCN), have achieved remarkable success by exploiting the spatial redundancy of neighboring pixels through convolutional processing. Though computationally efficient, we point out that such approaches are not statistically efficient during learning precisely because spatial redundancy limits the information learned from neighboring pixels. We demonstrate that (1) stratified sampling allows us to add diversity during batch updates and (2) sampled multi-scale features allow us to explore more nonlinear predictors (multiple fully-connected layers followed by ReLU) that improve overall accuracy. Finally, our objective is to show how a architecture can get performance better than (or comparable to) the architectures designed for a particular task. Interestingly, our single architecture produces state-of-the-art results for semantic segmentation on PASCAL-Context, surface normal estimation~\cite{Bansal16} on NYUDv2 dataset, and edge detection on BSDS without contextual post-processing.
\end{abstract}

\vspace{-1.5cm}

\section{Introduction}

\epigraph{Simplicity is the ultimate sophistication.} {\textit{Leonardo da Vinci}}

A surprising number of computer vision problems can be formulated as a dense pixel-wise prediction problem. These include {\bf low-level} tasks such as edge detection~\cite{dollar2013structured,martin2004learning,Xie15} and optical flow~\cite{baker2011database,DFIB15}, {\bf mid-level} tasks such as depth/normal recovery~\cite{Bansal16,Eigen15,eigen2014depth,saxena20083,Wang15}, and {\bf high-level} tasks such as keypoint prediction~\cite{GeorgiaBharathCVPR2014b,Ramanan2007}, object detection~\cite{huang2015densebox}, and semantic segmentation~\cite{chen2014semantic,farabet2013learning,Hariharan15,Long15,MostajabiYS15,shotton2006textonboost}.

Though such a formulation is attractive because of its generality, one obvious difficulty is the enormous associated output space. For example, a $100\times100$ image with $10$ discrete class labels per pixel yields an output label space of size $10^5$. One strategy is to treat this as a {\em spatially-invariant label prediction} problem, where one predicts a separate label per pixel using a convolutional architecture. Neural networks with convolutional output predictions, also called Fully Convolutional Networks (FCNs)~\cite{chen2014semantic,Long15,matan1991multi,Platt93}, appear to be a promising architecture in this direction.

But is this the ideal formulation of dense pixel-labeling? While {\em computationally efficient} for generating predictions at test time, we argue that it is {\em not statistically efficient} for gradient-based learning. Stochastic gradient descent (SGD) assumes that training data are sampled independently and from an identical distribution (\iid)~\cite{bottou2010large}. Indeed, a commonly-used heuristic to ensure approximately \iid samples is random permutation of the training data, which can significantly improve learnability~\cite{lecun2012efficient}. It is well known that pixels in a given image are highly correlated and not independent~\cite{hyvarinen2009natural}. Following this observation, one might be tempted to randomly permute pixels during learning, but this destroys the spatial regularity that convolutional architectures so cleverly exploit! 
In this paper, we explore the tradeoff between statistical and computational efficiency for convolutional learning, and investigate simply {\em sampling} a modest number of pixels across a small number of images for each SGD batch update, exploiting convolutional processing where possible.

\noindent\textbf{Contributions:} We experimentally validate that, thanks to spatial correlations between pixels, just sampling a small number of pixels per image is sufficient for learning. More importantly, sampling allows us to explore several avenues for improving both the efficiency and performance of FCN-based architectures.  
\begin{enumerate}
\item While most existing methods require up-sampling spatially-coarse predictions to the resolution of the original image pixel grid (\eg with deconvolution~\cite{Long15,Xie15} or interpolation~\cite{chen2014semantic}), sampling only requires on-demand computation of a sparse set of sampled features, therefore saving time and space during training (see Section~\ref{sec:approach}). 
\item The reduction in space and time allows us to explore more advanced architectures than prior work~\cite{Hariharan15,Long15}, which tend to use pixel-wise {\em linear} predictors defined over multi-scale ``hypercolumn'' features extracted from multiple layers of the network. Instead, we show that {\em nonlinear} predictors of hypercolumn features, implemented through multiple fully-connected layers followed by ReLU, significantly improve accuracy. We find a good tradeoff for learnability is convolutional processing for the lower-layers and on-demand sparse sampling of nonlinear pixel predictions.

\item In the case of skewed class label distribution, sampling offers the flexibility to let the model focus more on the rare classes. A good example is edge detection, where only $10\%$ of the ground truth are positive~\cite{Xie15}. Inspired by~\cite{girshick2014rcnn}, we demonstrate that a biased sample toward positives can greatly help the performance.

\item We show state-of-the-art results for edge detection on BSDS~\cite{amfm_pami2011}, out-performing the holistically-nested edge detection (HED) system of Xie et al.\ \cite{Xie15}. We also show competitive results for semantic segmentation on the PASCAL VOC-2012~\cite{Everingham10}, and more challenging PASCAL Context dataset where we achieve state of the art performance without contextual post processing~\cite{chen2014semantic}. Finally,~\cite{Bansal16} showed state-of-the-art performance for surface normal estimation using the same architecture.

\end{enumerate}

\section{Background}

In this section, we review related work by making use of a unified notation that will be used to describe our architecture. We address the pixel-wise prediction problem where, given an input image $X$, we seek to predict outputs $Y$.  
For pixel location $p$, the output can be binary $Y_p \in \{0,1\}$ (e.g., edge detection), multi-class $Y_p \in \{1,\dots,K\}$ (e.g., semantic segmentation), or real-valued $Y_p \in \mathbb{R}^N$ (e.g., surface normal prediction).  
There is rich prior art in modeling this prediction problem using hand-designed features (representative examples include~\cite{arbelaez2012semantic,carreira2012semantic,dollar2013structured,gould2009decomposing,liu2011nonparametric,munoz2010stacked,russell2009associative,shotton2006textonboost,tighe2010superparsing,tu2010auto,yao2012describing}).

{\bf Convolutional prediction:} We explore {\em spatially-invariant} predictors $f_{\theta,p}(X)$ that are end-to-end trainable over model parameters $\theta$. 
The family of fully-convolutional and skip networks~\cite{matan1991multi,Platt93} are illustrative examples that have been successfully applied to, e.g., edge detection~\cite{Xie15} and semantic segmentation~\cite{byeon2015scene,chen2014semantic,farabet2013learning,fischer2015flownet,Long15,liu2015parsenet,MostajabiYS15,noh2015learning,pinheiro2013recurrent}. Because such architectures still produce separate predictions for each pixel, numerous approaches have explored post-processing steps that enforce spatial consistency across labels via e.g., bilateral smoothing with fully-connected Gaussian CRFs~\cite{chen2014semantic,krahenbuhl2011efficient,zheng2015conditional} or bilateral solvers~\cite{Barron15}, dilated spatial convolutions~\cite{yu2015multi}, LSTMs~\cite{byeon2015scene}, and convolutional pseudo priors~\cite{xie2015convolutional}. In contrast, our work does {\em not} make use of such contextual post-processing, in an effort to see how far a pure ``pixel-level'' architecture can be pushed.

{\bf Multiscale features:} Higher convolutional layers are typically associated with larger receptive fields that capture high-level global context. Because such features may miss low-level details, numerous approaches have built predictors based on multiscale features extracted from multiple layers of a CNN~\cite{denton2015deep,Eigen15,eigen2014depth,farabet2013learning,pinheiro2013recurrent,Wang15}. Hariharan et al~\cite{Hariharan15} use the evocative term ``hypercolumns'' to refer to features extracted from multiple layers that correspond to the same pixel. Let $$h_p(X) = [c_1(p),c_2(p),\ldots, c_M(p)]$$ denote the multi-scale hypercolumn feature computed for pixel $p$, where $c_i(p)$ denotes the feature vector of convolutional responses from layer $i$ centered at pixel $p$ (and where we drop the explicit dependance on $X$ to reduce clutter).
Prior techniques for up-sampling include shift and stitch~\cite{Long15}, converting convolutional filters to dilation operations~\cite{chen2014semantic} (inspired by the {\em algorithme \`{a} trous}~\cite{Holschneider89}), and deconvolution/unpooling~\cite{fischer2015flownet,Long15,noh2015learning}. We similarly make use of multiscale features, but make use of {\em sparse} on-demand upsampling of filter responses, with the goal of reducing memory footprints during learning.

{\bf Pixel-prediction:} One may cast the pixel-wise prediction problem as operating over the hypercolumn features where, for pixel $p$, the final prediction is given by $$f_{\theta,p}(X) = g(h_p(X)).$$ We write $\theta$ to denote both parameters of the hypercolumn features $h$ and the pixel-wise predictor $g$. Training involves back-propagating gradients via SGD to update $\theta$. Prior work has explored different designs for $h$ and $g$. A dominant trend is defining a linear predictor on hypercolumn features, e.g., $g = w \cdot h_p$. FCNs~\cite{Long15} point out that linear prediction can be efficiently implemented in a coarse-to-fine manner by upsampling coarse predictions (with deconvolution) rather than upsampling coarse features. 
DeepLab~\cite{chen2014semantic} incorporates filter dilation and applies similar deconvolution and linear-weighted fusion, in addition to reducing the dimensionality of the fully-connected layers to reduce memory footprint. 
ParseNet~\cite{liu2015parsenet} added spatial context for a layer's responses by average pooling the feature responses, followed by normalization and concatenation. 
HED~\cite{Xie15} output edge predictions from intermediates layers, which are deeply supervised, and fuses the predictions by linear weighting.
Importantly, ~\cite{MostajabiYS15} and \cite{farabet2013learning} are noteable exceptions to the linear trend in that {\em non-linear} predictors $g$ are used. This does pose difficulties during learning -  \cite{MostajabiYS15} precomputes and stores superpixel feature map due to memory constraints, and so cannot be trained end-to-end. Our work demonstrates that sparse sampling of hypercolumn features allows for exploration of highly nonlinear $g$, which in turn significantly boosts performance.

{\bf Accelerating SGD:} There exists a large literature on accelerating stochastic gradient descent. We refer the reader to~\cite{bottou2010large} for an excellent introduction. Though naturally a sequential algorithm that processes one data example at a time, much recent work focuses on mini-batch methods that can exploit parallelism in GPU architectures~\cite{dean2012large} or clusters~\cite{dean2012large}. One general theme is efficient online approximation of second-order methods~\cite{bordes2009sgd}, which can model correlations between input features. Batch normalization~\cite{ioffe2015batch} computes correlation statistics between samples in a batch, producing noticeable improvements in convergence speed. Our work builds similar insights directly into convolutional networks without explicit second-order statistics.

\section{Approach}
\label{sec:approach}

This section describes our approach for pixel-wise prediction, making use of the notation introduced in the previous section. We first formalize our pixelwise prediction architecture, and then discuss statistically efficient mini-batch training.

\begin{figure*}[t]
\centering
\includegraphics[width=\linewidth]{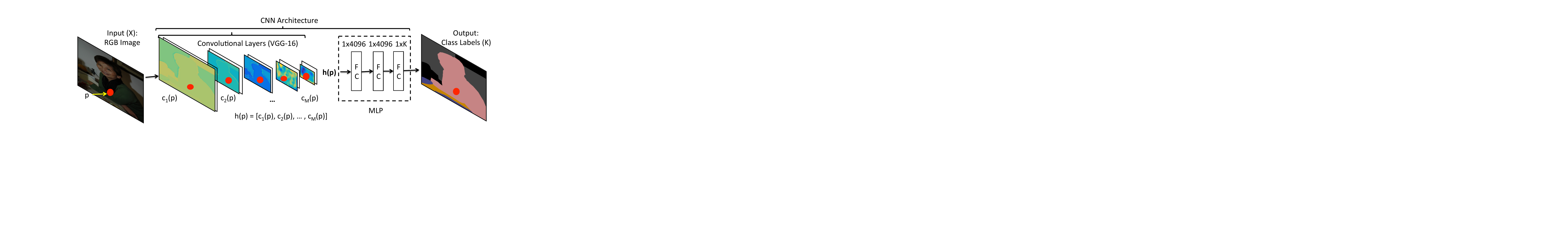}
\caption{Our PixelNet Architecture. Please see text for details.}
\label{fig:Net}
\end{figure*}

{\bf Architecture:} As in past work, our architecture makes use of multiscale convolutional features, which we write as a hypercolumn descriptor: $$h_p= [c_1(p),c_2(p),\ldots, c_M(p) ]$$ We learn a nonlinear predictor $f_{\theta,p} = g(h_p)$ implemented as a multi-layer perception (MLP)~\cite{bishop1995neural} defined over hypercolumn features. We use a MLP with ReLU activation functions, which can be implemented as a series of ``fully-connected'' layers. Importantly, the last layer must be of size $K$, the number of class labels or real valued outputs being predicted. We visualize our network in Figure~\ref{fig:Net}.

{\bf Dense predictions:}  We now describe an efficient method for generating dense pixel predictions with our network, which will be used at test-time. Dense prediction proceeds by (1) feedforward computation of convolutional responses at all layers $\{c_i\}$ and (2) bilinear interpolation (through ``deconvolution'')  of each response map to the original pixel resolution. This produces a dense grid of hypercolumn features, which are then (3) processed by pixel-wise MLPs implemented as 1x1 filters (representing each fully-connected layer). The memory intensive portion of this computation is the dense grid of hypercolumn features. This memory footprint is reasonable at test time because a single image can be processed at a time, but at train-time, we would like to train on batches containing many images as possible (to ensure diversity).

{\bf Sparse predictions:} We now describe an efficient method for generating sparse pixel predictions, which will be used at train-time (for efficient mini-batch generation).
Assume that we are given an image $X$ and a sparse set of (sampled) pixel locations $\{p_j\}$. We efficiently generate a sparse set of predictions at those pixels $\{f_{\theta,p_j} \}$ as follows:
we follow step (1) from above, but replace (2) with a sparse on-demand computation of hypercolumn features vectors at positions $\{ h_{p_j} \}$. To compute this set, we introduce a new {\em multi-scale sampling layer} (in \textit{Caffe}~\cite{jia2014caffe}) that directly extracts the 4 convolutional features corresponding to the 4 discrete locations in $c_i$ closest to pixel position $p_j$, and  then computes $c_i(p_j)$ via bilinear interpolation ``on the fly''.  This avoids the computation of a dense grid of hypercolumn features. Finally, step (3) can be implemented as a simple matrix-vector multiplication (by re-arranging the set of hypercolumn vectors $\{ h_{p_j} \}$ into a matrix). We experimentally demonstrate that this approach offers an excellent tradeoff between amortized computation and reduced storage, given that a modest number of pixels are sampled per image. If the number of samples is very small (`1' in the extreme case), one can further reduce computation with sparse convolutions (implemented say, by cropping the input image around the sample).
Finally, we note that our multi-scale sampling layers simply acts as a selection operation, for which a (sub) gradient can easily be defined. This means that backprop can also take advantage of sparse computations for nonlinear MLP layers and convolutional processing for the lower layers.

{\bf Mini-batch sampling:}
At each iteration of SGD training, the true gradient over the model parameters $\theta$ is approximated by computing the gradient over a relatively small set of samples from the training set. 
Approaches based on FCN~\cite{Long15} include features for all pixels from an image in a mini-batch.  As nearby pixels in an image are highly correlated, sampling them may potentially hurt learning. For instance, correlated samples may overfit to earlier images and require the use of lower learning rates, which slows convergence. To ensure a diverse set of pixels (while still enjoying the amortized benefits of convolutional processing), we settled on the following strategy: rather than using all pixels from a single image, we use a modest number of pixels (${\sim}2,000$) per image, but sample many images per batch. Naive computation of dense grid of hypercolumn descriptors takes almost all of the (GPU) memory, while $2,000$ samples takes a small amount using our sparse sampling layer. This allows us to explore more images per batch, significantly increasing sample diversity (as our experiments show, Sec.~\ref{sec:exp}). We explore the precise tradeoff between sampling size, number of images, and overall batch size in our experiments.

One might be tempted to think about naive ``straight-forward'' ways of sub-sampling with the existing architectures. One easy way to sub-sampling is to simply mask out pixel-level outputs. Naively computing a dense grid of hypercolumn descriptors and processing them with a nonlinear MLP would take more than \textbf{20X} memory compared to our approach. Slightly better would be masking the hypercolumn descriptors before MLP processing, which is still \textbf{16X} more expensive. We believe such ``implementation details'' are crucial for large-scale learning in today's world of SGD-based CNN optimization (c.f. batch normalization~\cite{ioffe2015batch}, residual learning~\cite{He2015}, etc).

{\bf Comparison with prior art:}
Unlike previous approaches (such as hypercolumns~\cite{Hariharan15} and FCN~\cite{Long15}), our approach sub-samples hypercolumn features from convolutional layers \textit{without} any up-sampling. Sub-sampling allows for the use of nonlinear functions (MLP) on such multiscale features, which in turn makes the architecture more generic (eliminating the need for task-specific normalization, scaling, or hand-tuning). As evidence, we use the same settings for three completely different problems (semantic sgmentation, surface normal estimation, and edge detection). For contrast, Xie and Tu~\cite{Xie15}  required significant modifications (such as deep supervision) to make FCNs applicable for low-level edge detection.

Long et al.\ ~\cite{Long15} argued against sampling and showed how the convergence is slowed when sampling few pixels.  While they experiment with $25-50$\% sampled pixels, we sample only $2$\% of total pixels in an image. We observed the similar behaviour when using a linear predictor (See Table~\ref{tb:ans1} for more details) but this issue of convergence goes away with the use of MLP. Not only the convergence, a linear predictor may require normalization/scaling, and careful hand-tuning for different tasks (as done in~\cite{Hariharan15,Long15}) as features across different convolutional layers lie in different dynamic ranges. On the contrary, our nonlinear MLP can learn to automatically take care of such issues.

\section{Experiments\label{sec:exp}}

In this section we describe our experimental evaluation. We apply our architecture (with minor modifications) to the high-level task of semantic segmentation, and the low-level task of edge detection. We show state-of-the-art\footnote{We briefly present the results of surface normal estimation here in this paper. Refer to~\cite{Bansal16} for more details.} results on PASCAL-Context~\cite{mottaghi_cvpr14} (without requiring contextual post-processing), competitive performance on PASCAL VOC 2012~\cite{Everingham10}, and advance the state of the art on the BSDS benchmark~\cite{amfm_pami2011}. We also perform a diagnostic evaluation of the effect of sampling and other hyperparameters/design choices.

{\bf Default network:} As with other methods~\cite{chen2014semantic,Long15,Xie15}, we fine-tune a VGG-16 network~\cite{SimonyanZ14a}. VGG-16 has 13 convolutional layers and three fully-connected ({\em fc}) layers. The convolutional layers are denoted as \{$1_1$, $1_2$, $2_1$, $2_2$, $3_1$, $3_2$, $3_3$, $4_1$, $4_2$, $4_3$, $5_1$, $5_2$, $5_3$\}. Following~\cite{Long15}, we transform the last two \emph{fc} layers to convolutional filters\footnote{For alignment purposes, we made a small change by adding a spatial padding of 3 cells for the convolutional counterpart of \emph{fc6} since the kernel size is $7\times 7$.}, 
and add them to the set of convolutional features that can be aggregated into our multi-scale hypercolumn descriptor. To avoid confusion with the {\bf fc} layers in our MLP, we will henceforth denote the {\em fc} layers of VGG-16 as conv-$6$ and conv-$7$. We use the following network architecture (unless otherwise specified): we extract hypercolumn features from conv-\{$1_2$, $2_2$, $3_3$, $4_3$, $5_3$\} with on-demand interpolation. 
We define a MLP over hypercolumn features with 3 fully-connected (fc) layers of size $4,096$ followed by ReLU~\cite{krizhevsky2012imagenet} activations, where the last layer outputs predictions for $K$ classes (with a soft-max/cross-entropy loss).

\begin{figure*}
\centering
\includegraphics[width=1\linewidth]{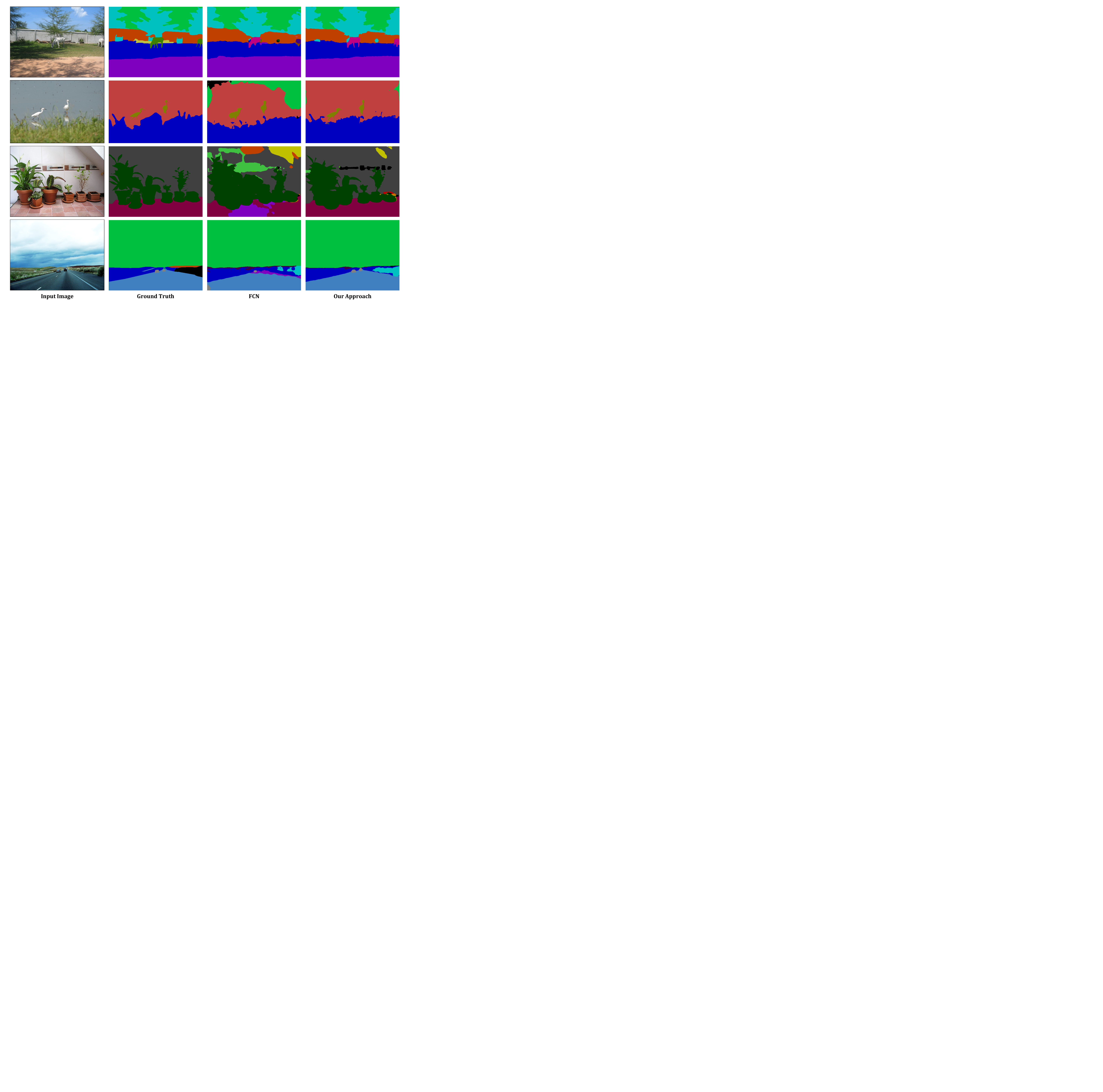}
\caption{Segmentation results on PASCAL-Context 59-class. Our approach uses an MLP to integrate information from both lower (\eg $1_2$) and higher (\eg \emph{conv-7}) layers, which allows us to better capture both global structure (object/scene layout) and fine details (small objects) compared to \emph{FCN-8s}.}
\label{fig:segQ}
\vspace{-0.5cm}
\end{figure*}

{\bf Default training:} For all the experiments we used the publicly available \emph{Caffe} library~\cite{jia2014caffe}. All trained models and code will be released. 
We make use of ImageNet-pretrained values for all convolutional layers, but train our MLP layers ``from scratch'' with Gaussian initialization ($\sigma=10^{-3}$) and dropout~\cite{srivastava2014dropout}. We fix momentum $0.9$ and weight decay $0.0005$ throughout the fine-tuning process. We use the following update schedule (unless otherwise specified): we tune the network for $80$ epochs with  a fixed learning rate ($10^{-3}$), reducing the rate by $10\times$ twice every 8 epochs until we reach $10^{-5}$. 

\subsection{Semantic Segmentation\label{sec:segres}}
\noindent {\bf Dataset.} The PASCAL-Context dataset~\cite{amfm_pami2011} augments the original sparse set of PASCAL VOC 2010 segmentation annotations~\cite{Everingham10} (defined for 20 categories) to pixel labels for the whole scene. While this requires more than $400$ categories, we followed standard protocol and evaluate on the 59-class and 33-class subsets. Though all the analysis in this paper are shown on PASCAL Context dataset~\cite{amfm_pami2011}, we also evaluated our approach on the standard PASCAL VOC-2012 dataset~\cite{Everingham10} to compare with a wide variety of approaches.

\noindent {\bf Qualitative Results.} 
We show qualitative outputs in Figure~\ref{fig:segQ} and compare against FCN-8s~\cite{Long15}.  Notice that we capture fine-scale details, such as the leg of birds (row 2) and plant leaves (row 3). 

\begin{table}[!t]
\small{
\centering
\begin{tabular}{ l c c } 
\toprule
\textbf{Settings} & $AC$ (\%) & $IU$ (\%) \\
\midrule
baseline (\emph{fc-3}, \emph{d-4096}) & 44.0 & 34.9 \\
\midrule
\emph{fc-1}  & 2.8 & 0.7 \\
\emph{fc-2}  & 1.7 & 0.1 \\
\midrule
\emph{d-1024} & 41.6 & 33.2 \\
\emph{d-2048} & 43.2 & 34.2 \\
\emph{d-6144} & 44.2 & 35.1 \\
\bottomrule
\end{tabular}
\vspace{0.2cm}
\caption{
Varying the number and dimension of the MLP \emph{fc} layers on the PASCAL Context 59-class segmentation task. Please see the text for detailed explanation of each setting.
}
\label{tb:ans1}}
\vspace{-0.5cm}
\end{table}

\begin{table}[!t]
\small{
\centering
\begin{tabular}{ l  c c } 
\toprule
\textbf{Settings} & $AC$ (\%) & $IU$ (\%) \\
\midrule
baseline (\emph{2000 $\times$ 5}) & 44.0 & 34.9 \\
\midrule
\emph{\ \ 500 $\times$ 5} & 43.7 & 34.8 \\
\emph{1000 $\times$ 5} & 43.8 & 34.7 \\
\emph{4000 $\times$ 5} & 43.9 & 34.9 \\
\midrule
\emph{\ \ 2000 $\times$ 1} & 32.6 & 24.6 \\
\emph{10000 $\times$ 1} & 33.3 & 25.2 \\
\bottomrule
\end{tabular}
\vspace{0.2cm}
\caption{Varying SGD mini-batch construction on the PASCAL Context 59-class segmentation task. \emph{N $\times$ M} refers to a mini-batch constructed from \emph{N} pixels sampled from each of M images (a total of N$\times$M pixels sampled for optimization). We see that a small number of pixels per image (500, or 2\%) are sufficient for learning. Put in another terms, given a fixed budget of N pixels per mini-batch, performance is maximized when spreading them across a large number of images M. This validates our central thesis that statistical diversity trumps the computational savings of convolutional processing during learning.}
\label{tb:ans2}}
\vspace{-0.5cm}
\end{table}

\noindent {\bf Evaluation Metrics.} We report results on the standard metrics of pixel accuracy ($AC$) and region intersection over union ($IU$) averaged over classes (higher is better). Both are calculated with DeepLab evaluation tools\footnote{\url{https://bitbucket.org/deeplab/deeplab-public/}}.

\noindent {\bf Analysis-1: Number of MLP \emph{fc} Layers.} We evaluate performance as a function of the number of MLP \emph{fc} layers. Our baseline system has two $4,096$-dimensional hidden layers (\emph{fc-3}).  We consider a linear predictor (\emph{fc-1}) (implemented as a single layer) and a single $4,096$-dimensional hidden layer (\emph{fc-2}). Most existing architectures combining different \emph{conv} layers~\cite{Hariharan15,Long15} are equivalent to a linear model (\emph{fc-1}), while networks that operate on modified features (\eg normalization~\cite{liu2015parsenet}, rescaling~\cite{bell2015inside}) can be viewed as employing a single (designed) intermediate layer.

We found it difficult to ensure convergence for single-layer predictors with the initial learning rate of $10^{-3}$, so we reduced it to $10^{-7}$. The results of the networks using the 59-class setup can be found in Table~\ref{tb:ans1} (middle rows). {\em Everything else is kept identical during the fine-tuning process}. The results are striking - models trained with fewer than 3 {\em fc} layers perform quite poorly: \emph{fc-2} constantly predicts the biggest class (``sky'') as the class label, while \emph{fc-1} behaves similarly, with some additional ``background'' and ``person'' pixels. This is consistent with ~\cite{Long15}'s observation that random sampling of patches during training can slow convergence. We posit that such careful initialization and training schemes (like stage-wise training~\cite{Long15}, $\ell_2$ normalization~\cite{liu2015parsenet} or deep supervision~\cite{Xie15}) are needed to train such networks. It is suprising that simply adding two hidden \emph{fc} layers appears to significantly simplify training.
Past work~\cite{liu2015parsenet} argues that convolutional features from different layers should be normalized before concatenation. We posit that two hidden {\em fc} layers can learn such normalizations automatically, though further investigation is needed.

\noindent {\bf Analysis-2: Dimension of MLP {\em fc} Layers.} Here we analyze performance as a function of the size of the MLP \emph{fc} layers. We experimented the following dimensions for our {\em fc} layers: $1,024$, $2,048$, $4,096$ (baseline) and $6,144$. Table~\ref{tb:ans1} (left, bottom rows) lists the results. We can see that with more dimensions the network tends to learn better, potentially because it can capture more information (and with drop-out alleviating over-fitting~\cite{srivastava2014dropout}). In the following experiments we fix the size to $4,096$, a good trade-off between performance and speed. 

\begin{table*}
\small{
\begin{center}
\begin{tabular}{ l  c c  c c } 
\toprule
\multirow{2}{*}{\textbf{Model}} & \multicolumn{2}{c}{\textbf{59-class}} & \multicolumn{2}{c}{\textbf{33-class}} \\
& $AC$ (\%) & $IU$ (\%) & $AC$ (\%) & $IU$ (\%) \\
\midrule
FCN-8s~\cite{DBLP:journals/corr/LongSD14} & 46.5 & 35.1 & 67.6 & 53.5 \\
FCN-8s~\cite{Long15} & 50.7 & 37.8 & - & - \\
DeepLab (v2~\cite{ChenPK0Y16})& - & 37.6 & - & - \\
\midrule
DeepLab (v2) + CRF~\cite{ChenPK0Y16} & - & 39.6 & - & - \\
CRF-RNN~\cite{zheng2015conditional} & - & 39.3 & - & - \\
ParseNet~\cite{liu2015parsenet} & - & 40.4 & - & - \\
ConvPP-8~\cite{xie2015convolutional} & - & \textbf{41.0} & - & - \\
\midrule
baseline (conv-\{$1_2$, $2_2$, $3_3$, $4_3$, $5_3$\}) & 44.0 & 34.9 & 62.5 & 51.1 \\
\midrule
conv-\{$1_2$, $2_2$, $3_3$, $4_3$, $5_3$, $7$\} ($0.25$,$0.5$) & 46.7 & 37.1 & 66.6 & 54.8 \\
conv-\{$1_2$, $2_2$, $3_3$, $4_3$, $5_3$, $7$\} ($0.5$) & 47.5 & 37.4 & 66.3 & 54.0 \\
conv-\{$1_2$, $2_2$, $3_3$, $4_3$, $5_3$, $7$\} ($0.5$-$1.0$) & 48.1 & 37.6 & 67.3 & 54.5 \\
conv-\{$1_2$, $2_2$, $3_3$, $4_3$, $5_3$, $7$\} ($0.5$-$0.25$,$0.5$,$1.0$) & \textbf{51.5} & \textbf{41.4} & \textbf{69.5} & \textbf{56.9} \\
\bottomrule
\end{tabular}
\vspace{0.2cm}
\caption{Our final results and baseline comparison on PASCAL-Context. Note that while most recent approaches spatial context post-processing~\cite{ChenPK0Y16,liu2015parsenet,xie2015convolutional,zheng2015conditional}, we focus on the FCN~\cite{Long15} per-pixel predictor as most approaches are its descendants. Also, note that we (without any CRF) achieve results better than previous approaches. CRF post-processing could be applied to any local unary classifier (including our method). Here we wanted to compare with other local models for a ``pure'' analysis.}
\label{tb:segperf}
\end{center}}
\end{table*}

\noindent {\bf Analysis-3: Number of Mini-batch Samples.} One of the critical questions regarding random sampling is the number of required sample. We plot performance as a function of the number of sampled pixels per image. In the first sampling experiment, we fix the batch size to $5$ images and sample $500$, $1000$, $2000$ (baseline) and $4000$ pixels from each image. The results are shown in Table~\ref{tb:ans2} (middle rows). We observe that: 1) even sampling only $500$ pixels per image (on average 2\% of the ${\sim}20,000$ pixels in an image) produces reasonable performance after just $96$ epochs. 2) performance is roughly constant as we increase the number of samples. 

We now perform experiments where the samples are drawn from the same image. When sampling $2000$ pixels from a single image (comparable in size to batch of $500$ pixels sampled from 5 images), performance dramatically drops. This phenomena consistently holds for additional pixels (Table~\ref{tb:ans2}, bottom rows), verifying our central thesis that statistical diversity of samples can trump the computational savings of convolutional processing during learning.

\noindent {\bf Adding conv-$7$.} While our diagnostics reveal the importance of architecture design and sampling, our best results still do not quite reach the state-of-the-art. For example, a single-scale FCN-32s~\cite{Long15}, without any low-level layers, can already achieve $35.1$.  This suggests that their penultimate \emph{conv-7} layer does capture cues relevant for pixel-level prediction. In practice, we find that simply concatenating {\emph{conv-7}} significantly improves performance.

\begin{table*}
\small{
\begin{center}
\begin{tabular}{ l c c c c c c } 
\toprule
Model & features (\#) & \emph{fc} (\#) & sample (\#) & Memory (MB) & Size (MB) & $BPS$ \\
\midrule
FCN-32s~\cite{Long15} & 4,096 & 1 & 50,176 & 2,056 & 570 & 20.0 \\
FCN-8s~\cite{Long15} & 4,864 & 1 & 50,176 & 2,010 & 518 & 19.5 \\
\midrule
FCN, conv-\{$1_2$, $3_3$, $5_3$\}, \emph{fc-1} & 1,056 & 1 & 50,176 & 2,267 & 1,150 & 6.5 \\
FCN, conv-\{$1_2$, $3_3$, $5_3$\}, \emph{fc-2} & 1,056 & 2 & 50,176 & 3,066 & 1,165 & 4.2 \\
FCN, conv-\{$1_2$, $3_3$, $5_3$\}, \emph{fc-3} & 1,056 & 3 & 50,176 & 3,914 & 1,232 & 1.4 \\
\midrule
FCN, conv-\{$1_2$, $3_3$, $5_3$\}, \emph{fc-1} & 1,056 & 1 & 2,000 & 2,092 & 1,150 & 5.5 \\
FCN, conv-\{$1_2$, $3_3$, $5_3$\}, \emph{fc-2} & 1,056 & 2 & 2,000 & 2,138 & 1,165 & 5.4 \\
FCN, conv-\{$1_2$, $3_3$, $5_3$\}, \emph{fc-3} & 1,056 & 3 & 2,000 & 2,234 & 1,232 & 5.1 \\
\midrule
Ours, conv-\{$1_2$, $3_3$, $5_3$\}, \emph{fc-1} & 1,056 & 1 & 2,000 & 322 & 60 & 43.3 \\
Ours, conv-\{$1_2$, $3_3$, $5_3$\}, \emph{fc-2} & 1,056 & 2 & 2,000 & 368 & 74 & 38.7 \\
Ours, conv-\{$1_2$, $3_3$, $5_3$\}, \emph{fc-3} & 1,056 & 3 & 2,000 & 465 & 144 & 24.5 \\
\midrule
Ours, conv-\{$1_2$, $3_3$, $5_3$, $7$\}, \emph{fc-3} & 5,152 & 3 & 2,000 & 1,024 & 686 & 8.8 \\
\bottomrule
\end{tabular}
\caption{Efficiency/performance comparison between several models. We record the number of dimensions for hypercolumn features, number of \emph{fc} layers on the top, number of samples (for our model), memory usage, model size, number of mini-batch updates per second ($BPS$ measured by forward/backward passes). We use a single $224\times 224$ image as the input, and additional \emph{fc} layers are all of $4,096$ dimentions. The speed testing is done on Titan-X averaged over $10$ iterations. We compared our network with FCN~\cite{Long15} where a deconvolution layer is used to upsample the result in various settings. Note that besides FCN-8s and FCN-32s here we first compute the upsampled feature map, then apply the classifiers for FCN~\cite{Long15} due to the additional \emph{fc} layers. This is necessary for MLPs with more \emph{fc} layers. Even though our sampling layer is currently implemented in CPU, it still outperforms \emph{deconv} layers in both speed and memory/hard-disk usage. We also tried to include \emph{conv-7} for \emph{deconv} but the blob size goes beyond INT\_MAX.}
\label{tb:efficiency}
\end{center}}
\end{table*}

Following the same training process, the results of our model with \emph{conv-7} features are shown in Table~\ref{tb:segperf}. From this we can see that \emph{conv-7} is greatly helping the performance of semantic segmentation. Even with reduced scale, we are able to obtain a similar $IU$ achieved by FCN-8s~\cite{Long15}, without any extra modeling of context~\cite{chen2014semantic,liu2015parsenet,xie2015convolutional,zheng2015conditional}. For fair comparison, we also experimented with single scale training with 1) half scale $0.5\times$, and 2) full scale $1.0\times$ images. We find the results are better without $0.25\times$ training, reaching $37.4\%$ and $37.6\%$ $IU$, respectively, even closer to the FCN-8s performance ($37.8\%$ $IU$). For the 33-class setting, we are already doing better with the baseline model plus \emph{conv-7}.

\noindent {\bf Analysis-4: Multi-scale.} All previous experiments process test images at a single scale ($0.25\times$ or $0.5\times$ its original size), whereas most prior work~\cite{chen2014semantic,liu2015parsenet,Long15,zheng2015conditional} use multiple scales from full-resolution images. A smaller scale allows the model to access more context when making a prediction, but this can hurt performance on small objects. Following past work, we explore test-time averaging of predictions across multiple scales. We tested combinations of $0.25\times$, $0.5\times$ and $1\times$. For efficiency, we just fine-tune the model trained on small scales (right before reducing the learning rate for the first time) with an initial learning rate of $10^{-3}$ and step size of $8$ epochs, and end training after $24$ epochs. The results are also reported in Table~\ref{tb:segperf}. Multi-scale prediction generalizes much better ($41.0\%$ $IU$). Note our pixel-wise predictions do not make use of contextual post-processing (even outperforming some methods that post-processes FCNs to do so~\cite{ChenPK0Y16,zheng2015conditional}).

\noindent {\bf Efficiency.} We compared our speed, model size, and memory usage of our network to FCN~\cite{Long15} (same architecture) in Table~\ref{tb:efficiency}. Removing the deconvolution layer reduces memory consumption. 

\noindent {\bf PASCAL VOC-2012. } Finally we use the same settings and evaluate our approach on PASCAL VOC-2012. Our approach achieves mAP of \textbf{69.7\%}\footnote{Per-class performance is available at \url{http://host.robots.ox.ac.uk:8080/anonymous/PZH9WH.html}.}. This is much better than previous approaches, e.g. 62.7\% for Hypercolumns~\cite{Hariharan15}, 62\% for FCN~\cite{Long15}, ~67\% for DeepLab (without CRF)~\cite{chen2014semantic} etc. Our performance on VOC-2012 is similar to Mostajabi et al~\cite{MostajabiYS15} despite the fact we use information from only 6 layers while they used information from all the layers. In addition, they use a rectangular region of 256$\times$256 (called \textit{sub-scene}) around the super-pixels. We posit that fine-tuning (or back-propagating gradients to conv-layers) enables efficient and better learning with even lesser layers, and without extra \textit{sub-scene} information in an end-to-end framework. Finally, the use of super-pixels in~\cite{MostajabiYS15} inhibit capturing detailed segmentation mask (and rather gives ``blobby'' output), and it is computationally less-tractable to use their approach for per-pixel optimization as information for each pixel would be required to be stored on disk.

\subsection{Surface Normal Estimation\label{sec:surf_normals}}
PixelNet architecture was first proposed in our work~\cite{Bansal16} on 2D-to-3D model alignment via surface normal estimation. Here we extract some of the results from~\cite{Bansal16} to show the effectiveness of this architecture for the mid-level task of surface normal estimation. The NYU Depth v2 dataset~\cite{Silberman12} is used to evaluate the surface normal maps. The criteria introduced by Fouhey et al.~\cite{Fouhey13a} is used to compare our approach~\cite{Bansal16} against prior work~\cite{Eigen15,Fouhey13a}. Six statistics are computed over the angular error between the predicted normals and depth-based normals -- \textbf{Mean}, \textbf{Median}, \textbf{RMSE}, \textbf{11.25$^\circ$}, \textbf{22.5$^\circ$}, and \textbf{30$^\circ$} -- using the normals of Ladicky et al.\ ~\cite{Ladicky14} as ground truth (Note that these normals are computed from depth data obtained using kinect). The first three criteria capture the mean, median, and RMSE of angular error, where lower is better. The last three criteria capture the percentage of pixels within a given angular error, where higher is better. Table~\ref{tab:nyud2_scene} compares our approach~\cite{Bansal16} with previous state-of-the-art approaches. Please refer to~\cite{Bansal16} for more details on surface normal estimation. 

Unlike the task of semantic segmentation and edge detection, we use a single scale for estimating surface normal maps. We will release the results of using multi-scale approach for surface normal estimation in a future version.

\begin{table}[t]
\small{
\setlength{\tabcolsep}{3pt}
\def\arraystretch{1.2}
\center
\begin{tabular}{@{}l c c c c c c }
\toprule
\textbf{NYUDv2 test}  & Mean  &   Median & RMSE &  11.25$^\circ$ & 22.5$^\circ$ &  30$^\circ$ \\
\midrule
Fouhey et al.~\cite{Fouhey13a}	      &  35.3     &	31.2	 &   41.4	 &   16.4      &	  36.6   &	48.2	\\ 
E-F (AlexNet)~\cite{Eigen15} 	      & 23.7      &	15.5	 &    -	 &   39.2     & 62.0	     &	71.1	\\ 
E-F (VGG-16)~\cite{Eigen15} 	      & 20.9      &	13.2	 &    -	 &   44.4     & 67.2	     &	75.9	\\ 
\midrule
Ours~\cite{Bansal16}		      &	  \textbf{19.8}     &	\textbf{12.0}	 &   \textbf{28.2}	 &   \textbf{47.9}     &	   \textbf{70.0}  & \textbf{77.8}		\\  	
\bottomrule
\end{tabular}
\vspace{-3pt}
\caption{NYUv2 surface normal prediction from~\cite{Bansal16}.}
\label{tab:nyud2_scene}
}
\vspace{-0.1cm}
\end{table}

\subsection{Edge Detection\label{sec:edgeres}}
In this section, we demonstrate that our same architecture can produce state-of-the-art results for low-level edge detection. The standard dataset for edge detection is BSDS-500~\cite{amfm_pami2011}, which consists of $200$ training, $100$ validation, and $200$ testing images. Each image is annotated by ${\sim}5$ humans to mark out the contours. We use the same augmented data (rotation, flipping, totaling $9600$ images without resizing) used to train the state-of-the-art Holistically-nested edge detector (HED)~\cite{Xie15}. We report numbers on the testing images. During training, we follow HED and only use positive labels where a consensus ($\ge 3$ out of $5$) of humans agreed.

\begin{table}
\small{
\begin{center}
\begin{tabular}{ l c c c  } 
\toprule
 \  & \textbf{ODS} & \textbf{OIS} & \textbf{AP} \\
\midrule
conv-\{$1_2$, $2_2$, $3_3$, $4_3$, $5_3$\} \emph{Uniform} & .767 & .786 & .800 \\
\midrule
conv-\{$1_2$, $2_2$, $3_3$, $4_3$, $5_3$\} (25\%) & .792 & .808 & .826 \\
conv-\{$1_2$, $2_2$, $3_3$, $4_3$, $5_3$\} (50\%) & .791 & .807 & .823 \\
conv-\{$1_2$, $2_2$, $3_3$, $4_3$, $5_3$\} (75\%) & .790 & .805 & .818 \\
\bottomrule
\end{tabular}
\vspace{0.5cm}
\caption{Comparison of different sampling strategies during training. \emph{Top row:} Uniform pixel sampling. \emph{Bottom rows:} Biased sampling of positive examples. We sample a fixed percentage of positive examples ($25\%$,$50\%$ and $75\%$) for each image. Notice a significance difference in performance.}
\label{tb:fgrate}
\end{center}}
\vspace{-0.5cm}
\end{table}

\begin{table}
\scriptsize{
\begin{center}
\begin{tabular}{ l c c c  }
\toprule
 \  & \textbf{ODS} & \textbf{OIS} & \textbf{AP} \\
\midrule
Human~\cite{amfm_pami2011} & .800 & .800 & - \\
\midrule
Canny & .600 & .640 & .580 \\
Felz-Hutt~\cite{felzenszwalb2004efficient} & .610 & .640 & .560 \\
\midrule
gPb-owt-ucm~\cite{amfm_pami2011} & .726 & .757 & .696 \\
Sketch Tokens~\cite{lim2013sketch} & .727 & .746 & .780 \\
SCG~\cite{xiaofeng2012discriminatively} & .739 & .758 & .773 \\
\midrule
PMI~\cite{isola2014crisp} & .740 & .770 & .780 \\
\midrule
SE-Var~\cite{dollar2015fast} & .746 & .767 & .803 \\
OEF~\cite{hallman2015oriented} & .749 & .772 & .817 \\
\midrule
DeepNets~\cite{kivinen2014visual} & .738 & .759 & .758 \\
CSCNN~\cite{hwang2015pixel} & .756 & .775 & .798 \\
HED~\cite{Xie15} & .782 & .804 & .833 \\
HED~\cite{Xie15} (Updated version) & .790 & .808 & .811 \\
HED merging~\cite{Xie15} (Updated version) & .788 & .808 & {\bf .840} \\
\midrule
conv-\{$1_2$, $2_2$, $3_3$, $4_3$, $5_3$\} (50\%) & .791 & .807 & .823 \\
conv-\{$1_2$, $2_2$, $3_3$, $4_3$, $5_3$, $7$\} (50\%) & \textbf{.795} & \textbf{.811} & .830 \\
\midrule
conv-\{$1_2$, $2_2$, $3_3$, $4_3$, $5_3$\} (25\%)-($0.5{\times}$,$1.0{\times}$) & .792 & .808 & .826 \\
conv-\{$1_2$, $2_2$, $3_3$, $4_3$, $5_3$, $7$\} (25\%)-($0.5{\times}$,$1.0{\times}$) & \textbf{.795} & \textbf{.811} & .825 \\
\midrule
conv-\{$1_2$, $2_2$, $3_3$, $4_3$, $5_3$\} (50\%)-($0.5{\times}$,$1.0{\times}$) & .791 & .807 & .823 \\
conv-\{$1_2$, $2_2$, $3_3$, $4_3$, $5_3$, $7$\} (50\%)-($0.5{\times}$,$1.0{\times}$) & \textbf{.795} & \textbf{.811} & .830 \\
\midrule
conv-\{$1_2$, $2_2$, $3_3$, $4_3$, $5_3$, $7$\} (25\%)-($1.0{\times}$) & .792 & .808 & .837 \\
conv-\{$1_2$, $2_2$, $3_3$, $4_3$, $5_3$, $7$\} (50\%)-($1.0{\times}$) & .791 & .803 & \textbf{.840} \\
\bottomrule
\end{tabular}
\vspace{0.2cm}
\caption{Evaluation on BSDS~\cite{amfm_pami2011}. Our approach performs better than previous approaches \textit{specifically} trained for edge detection.}
\label{tb:bsds}
\end{center}}
\vspace{-0.5cm}
\end{table}

\noindent {\bf Baseline.} We use the same baseline network that was defined for semantic segmentation, only making use of pre-trained \emph{conv} layers. A sigmoid cross-entropy loss is used to determine the whether a pixel is belonging to an edge or not. Due to the highly skewed class distribution, we also normalized the gradients for positives and negatives in each batch (as in~\cite{Xie15}).

\noindent {\bf Training.} We use our previous training strategy, consisting of batches of $5$ images with a total sample size of $10,000$ pixels. Each image is randomly resized to half its scale (so $0.5$ and $1.0$ times) during learning. The initial learning rate is again set to $10^{-3}$. However, since the training data is already augmented, we found the network converges much faster than when training for segmentation. To avoid over-training and over-fitting, we reduce the learning rate at $15$ epochs and $20$ epochs (by a factor of $10$) and end training at $25$ epochs.

\noindent {\bf Baseline Results.} The results on BSDS, along with other concurrent methods, are reported in Table~\ref{tb:bsds}. We apply standard non-maximal suppression and thinning technique using the code provided by~\cite{dollar2013structured}. We evaluate the detection performance using three standard measures: fixed contour threshold (ODS), per-image best threshold (OIS), and average precision (AP). 

\begin{figure}
\centering
\includegraphics[width=\linewidth]{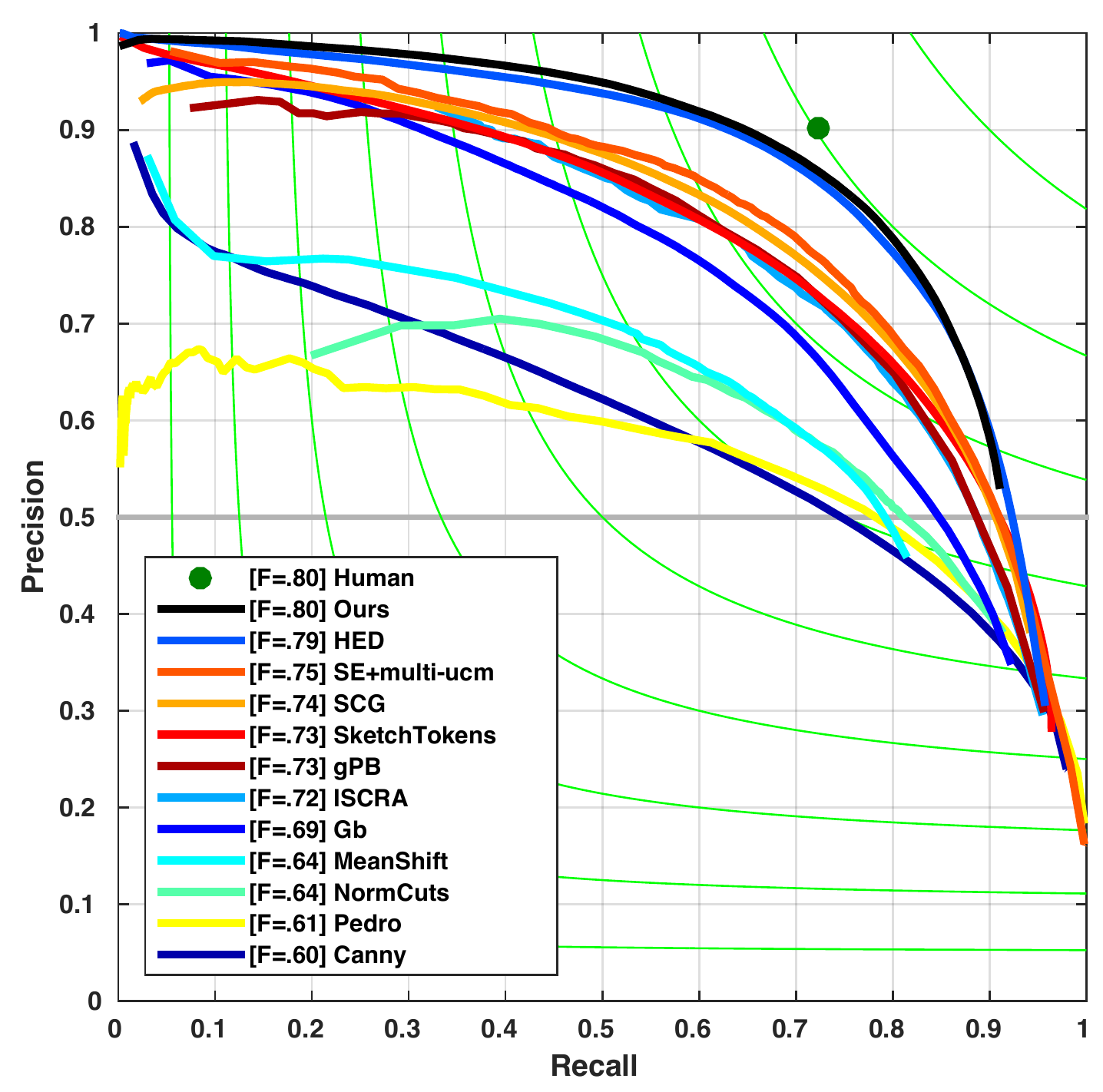}
\caption{Results on BSDS~\cite{amfm_pami2011}. While our curve is mostly overlapping with HED, our detector focuses on more high-level semantic edges. See qualitative results in Fig.\ref{fig:edges}. }
\label{fig:plot}
\end{figure}

\noindent {\bf Analysis-1: Sampling.} Whereas uniform sampling sufficed for semantic segmentation~\cite{Long15}, we found the extreme rarity of positive pixels in edge detection required focused sampling of positives. We compare different strategies for sampling a fixed number ($2000$ pixels per image) training examples in Table~\ref{tb:fgrate}. Two obvious approaches are uniform and balanced sampling with an equal ratio of positives and negatives (shown to be useful for object detection~\cite{girshick2015fast}). We tried ratios of $0.25$, $0.5$ and $0.75$, and found that balancing consistently improved performance\footnote{Note that simple class balancing~\cite{Xie15} in each batch is already used, so the performance gain is \emph{unlikely} from label re-balancing.}. 

\begin{figure*}
\centering
\includegraphics[width=1\linewidth]{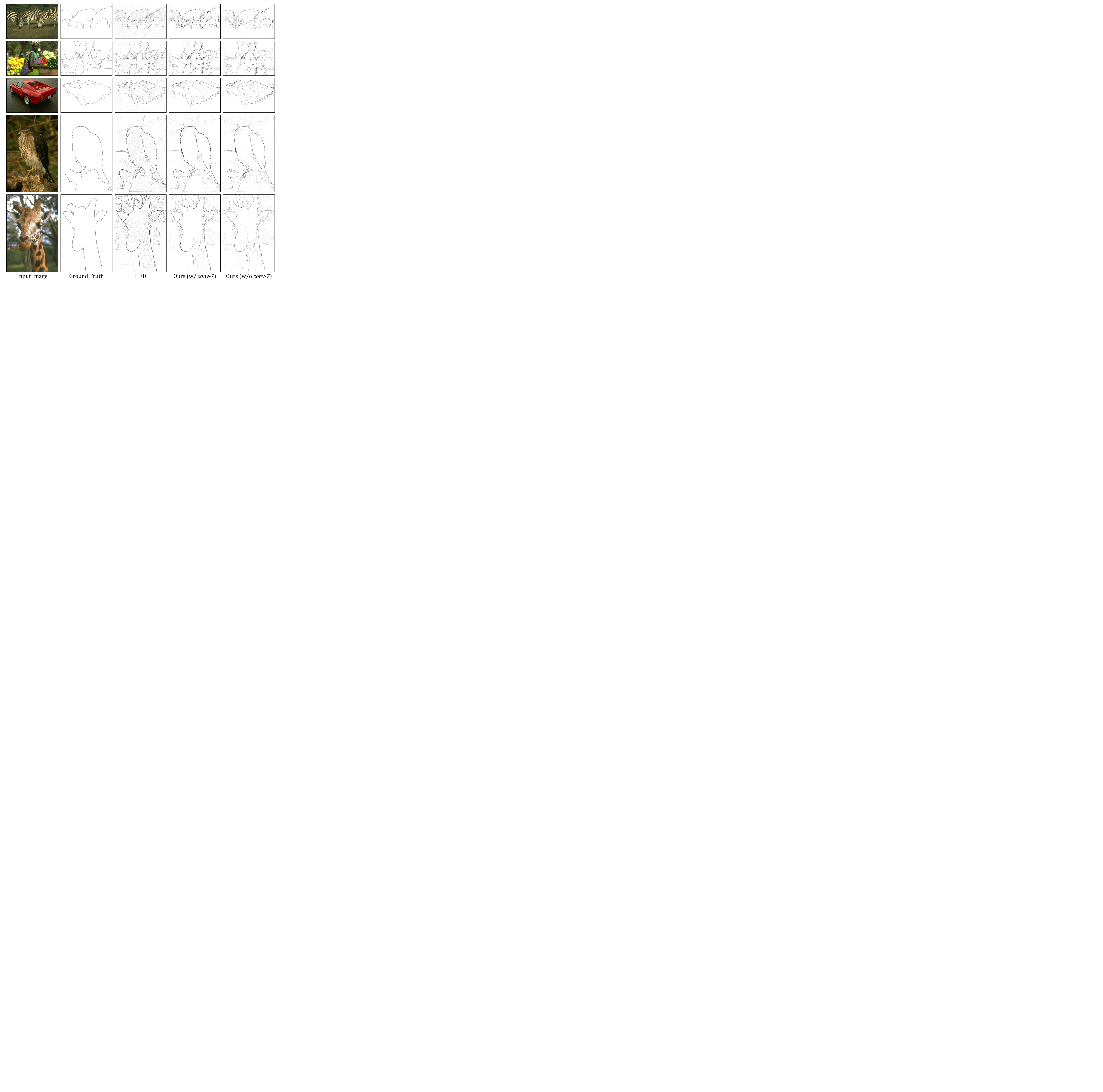}
\caption{Qualitative results for edge detection. Notice that our approach generates more semantic edges for \emph{zebra}, \emph{eagle}, and \emph{giraffe} compared to HED~\cite{Xie15}. Best viewed in the electronic version.}
\label{fig:edges}
\end{figure*}

\noindent {\bf Analysis-2: \emph{conv-7}.} We previously found that adding features from higher layers is helpful for semantic segmentation. Are such high-level features also helpful for edge detection, generally regarded as a low-level task? To answer this question, we again concatenated \emph{conv-7} features with other \emph{conv} layers \{ $1_2$, $2_2$, $3_3$, $4_3$, $5_3$ \}. Please refer to the results at Table~\ref{tb:bsds}, using the second sampling strategy. We find it still helps performance a bit, but not as significantly for semantic segmentation (clearly a high-level task). Our final results as a single output classifier are very competitive to the state-of-the-art.

Qualitatively, we find our network tends to have better results for semantic-contours (\eg around an object), particularly after including \emph{conv-7} features. Figure~\ref{fig:edges} shows some qualitative results comparing our network with the HED model. Interestingly, our model explicitly removed the edges inside the \emph{zebra}, but when the model cannot recognize it (\eg its head is out of the picture), it still marks the edges on the black-and-white stripes. Our model appears to be making use of much higher-level information than past work on edge detection. 

\section{Discussion} We have described a convolutional pixel-level architecture that, with minor modifications, produces state-of-the-art accuracy on diverse high-level, mid-level~\cite{Bansal16}, and low-level tasks. We demonstrate impressive results\footnote{We ran a vanilla version of our approach for depth estimation, and achieved near state-of-the-art performance (on NYU-v2 depth dataset) with a simple scale-invariant loss function~\cite{eigen2014depth}. We will add the results of depth estimation after more careful analysis in a later version.} on highly-benchmarked semantic segmentation, surface normal estimation~\cite{Bansal16}, and edge datasets. Our results are made possible by careful analysis of computational and statistical considerations associated convolutional predictors. Convolution exploits spatial redundancy of pixel neighborhoods for efficient computation, but this redundancy also impedes learning. We propose a simple solution based on stratified sampling that injects diversity while taking advantage of amortized convolutional processing. Finally, our efficient learning scheme allow us to explore nonlinear functions of multi-scale features that encode both high-level context and low-level spatial detail, which appears relevant for most pixel prediction tasks.

\vspace{0.1cm}
\scriptsize{\noindent\textbf{Acknowledgements:} This work was in part supported by NSF Grants IIS 0954083, IIS 1618903, and support from Google and Facebook. AB and XC would like to thank Abhinav Shrivastava and Saining Xie for useful discussion.}

{\small
\bibliographystyle{ieee}
\bibliography{shortstrings,references}
}

\end{document}